\documentclass[runningheads]{llncs}
\usepackage[T1]{fontenc}
\usepackage{hyperref}
\usepackage{graphicx}
\usepackage{xcolor}

\usepackage{color}

\usepackage{eurosym}
\usepackage{multirow}
\usepackage{soul}

\begin{document}
\title{Retrieval-Enhanced Real Estate Appraisal}
%
%
\author{Simon Popelier\inst{1}\orcidID{0009-0002-3469-9713} \and
Matthieu X. B. Sarazin\inst{1}\orcidID{0000-0001-8932-1430} \and
Maximilien Bohm\inst{2} \and Mathieu Gierski\inst{2} \and Hanna Mergui\inst{2} \and Matthieu Ospici\inst{1}\orcidID{0000-0002-7816-721X} \and Adrien Bernhardt\inst{1}\orcidID{0009-0004-8308-9457}}
\authorrunning{S. Popelier et al.}
%
\institute{Homiwoo, 10 rue de la Boétie, 75008 Paris, France \\
\url{https://www.homiwoo.com/} \\
\email{\{simon,matthieus,matthieuo,adrien\}@homiwoo.com}
\and
\'Ecole Polytechnique, Route de Saclay, 91128 Palaiseau, France \email{\{maximilien.bohm,mathieu.gierski,hanna.mergui\}@polytechnique.edu}
}
\maketitle              
\begin{abstract}
The Sales Comparison Approach (SCA) is one of the most popular when it comes to real estate appraisal. Used as a reference in real estate expertise and as one of the major types of Automatic Valuation Models (AVM), it recently gained popularity within machine learning methods. The performance of models able to use data represented as sets and graphs made it possible to adapt this methodology efficiently, yielding substantial results. SCA relies on taking past transactions (comparables) as references, selected according to their similarity with the target property's sale. 
In this study, we focus on the selection of these comparables for real estate appraisal. We demonstrate that the selection of comparables used in many state-of-the-art algorithms can be significantly improved by learning a selection policy instead of imposing it. Our method relies on a hybrid vector-geographical retrieval module capable of adapting to different datasets and optimized jointly with an estimation module. We further show that the use of carefully selected comparables makes it possible to build models that require fewer comparables and fewer parameters with performance close to state-of-the-art models. All our evaluations are made on five datasets which span areas in the United States, Brazil, and France.

\end{abstract}

\section{Introduction}
Real estate appraisal is a highly requested exercise by both general public and financial institutions alike. Market value, defined as the price that could be obtained from its sale on an open market \cite{hoeven_appraiser-based_2022}, is key to several operations. These include the management of portfolios, the granting of mortgage loans, the assessment of the viability of a real estate project, or the renting of a property. Evaluating the risk of these decisions is capital, requiring appraisals to respect standards of performance and consistency across vast areas comprising a diversity of markets. Moreover, market actors require explainable models to inspire confidence in downstream decision-making, or the ability to intuitively interpret mistakes otherwise.
Evaluating market values is an arduous task. A plethora of factors can influence the price of a property, ranging from physical attributes to its geographical location or the market situation at the time of sale. Solving this high-dimensional problem is difficult for human appraisers and remains challenging for computational methods. To this can be added the subjectivity and personal context of historical transactions, leading to an increased potential bias for statistical and machine learning techniques \cite{Birkeland2021}.

A variety of methods have been developed by experts and integrated into AVMs and machine learning systems to solve these problems. One standard current approach for expert appraising is through SCA, i.e. using similar and neighboring properties whose transaction values are known as references in order to interpolate the price of a new transaction. While traditional algorithms closely mimic this behavior through explicitly defined functions of spatial interpolation \cite{helbich2014,choi2015}, more recently deep learning models have been favored due to their ability to learn implicit spatially-complex local environments and treat sets of references together \cite{viana2021,Han2023,sellam2024boosting}.

Nonetheless, whereas these models have grown sophisticated, leveraging representation learning practices for price regression, their selection policy remains relatively naive. Most rely on temporal or geographical heuristics to gather references, e.g. selecting comparables using predefined representations with $k$-Nearest Neighbors (k-NN) logic \cite{Han2023}, yet none seem to learn their selection policy. 

The aim of our work is thus to propose a new approach to the selection of comparables by learning representations based on geographic and other features as well. As the selection of such comparables is essential for increasing the predictive quality of the models, advances in the selection of comparables is crucial for the field of real estate appraisal.

As such, we introduce the following technical contributions:

\begin{itemize}
\item We propose a new comparable selection framework based on retrieval-enhanced machine learning (REML) \footnote{\url{https://github.com/homiwoo/retrieval-enhanced-real-estate-appraisal}} ~\cite{zamani}. When isolating this comparable selection module, we demonstrate that learning to select comparables produces higher-quality comparables compared to traditional selection methods, as evidenced by the lower number of comparables necessary to obtain similar model performance. In turn, fewer comparables allows for easier human examination of retrieved properties and subsequent model predictions, increasing model explainability and confidence of third-party decision-makers. Finally, similar performance is obtained with up to 22$\times$ less parameters compared to state-of-the-art models \cite{viana2021}, further boosting explainability.
\item We evaluate our propositions across state-of-the-art models datasets and on a new dataset of French real estate transactions with explicit temporality.
\end{itemize}

\section{Related Works}
\subsection{Real estate appraisal}

Hedonic Appraisal techniques \cite{GOODMAN1998,valier2020} (dating back to 1939) first approached the problem directly by weighing each structural component \cite{wing2003} of the property as an inherently valuable asset. However, they are in truth relative to local contexts and thus require strong localization representation.
In comparison, SCA grounds the evaluation in a local context that provides explainability, resilience to concept drift~\cite{lu2019}, and partly bypasses the need for strong localization representation. Indeed, recent works focus on localization and neighborhood attributes, the former being often considered as one if not the major explanatory variable for real estate price~\cite{gao2022}.
The new wave of deep learning models has introduced sophisticated learning schemes, such as implementing a fine-tuning stage using contrastive learning specifically dedicated to universal representations \cite{du2023}. Encoding is rendered rich enough such that a Multi-Layer Perceptron (MLP) predictor can efficiently use it to predict prices. 

In these works, comparables are selected primarily with manually specified rules on the basis of geographical and raw euclidean feature distance, preventing finer-grained contextual comparable search. Our method overcomes these limitations through an innovative learned selection of comparables.

\subsection{Retrieval-enhanced machine learning}


External memory can be beneficial to machine learning algorithms to enhance their performance and ground their predictions in factual knowledge \cite{zhang2021}. In domains such as natural language processing \cite{khandelwal2020,guu2020} or vision \cite{chen2023reimagen,iscen2023}, efficient techniques have been developed to help models make informed decisions by retrieving elements from external memory. These techniques issued from the information retrieval (IR) literature fall under the category of retrieval-enhanced machine learning (REML) \cite{zamani}. Our approach adapts core REML ideas to complement modern appraisal approaches for both retrieval strategies and retriever training. 


\subsubsection{Retrieval strategies}
Traditional retrieval relied on domain-based heuristics to select elements from the database. They can provide an intuitive and explainable solution. However, their design is limited by the usual constraints of rule-based approaches. BM25 \cite{Amati2009} for text retrieval lacks context and semantics. In the context of real estate appraisal, the topical relevance provided by these approaches takes the form of spatial relevance. Modern retrieval-enhanced machine learning relies instead on dense vector search \cite{lee2019} or combines both traditional and vector-based approaches \cite{askari2023}. Our framework relies on this strategy and combines both geographical and vector information to find comparables. 




\subsubsection{Retriever training}
With the exception of tasks where large foundation models exist,  both the retriever and the downstream model need to be trained. They can be trained jointly \cite{izacard2022}, or they can be trained in distinct stages \cite{lin2024}. Training a retriever separately requires a dedicated training set, and some works argue in favor of using feedback from the downstream model. For instance, attention scores from the downstream model can be used to teach the retriever by distillation of attention \cite{izacard2022}. Attention scores can even be self-sufficient, allowing retriever and downstream models to fuse \cite{wu2022}. We take inspiration from this last approach in order to train a single model end-to-end in a single stage by optimizing a single objective function.

\section{Method}
\subsection{Overview}


We propose a Retrieval-Enhanced Appraisal module called (REA) able to select comparables by learning their representation. We use this module in an extended model (EREA) that can reach performance close to state-of-the-art while retaining high explainability. EREA and REA are described in Figure~\ref{fig1}. 



Training alternates between comparable selection by retrievers, their exploitation to predict a market value by a downstream model and the update of these two modules. We are using two retrieval mechanisms, one based on geographical proximity (geographic retrieval) and the other on vector similarity (vector retrieval). The downstream model encodes selected comparables and exploit their representation to provide a market value estimation. This vector representation is updated using gradient descent and fed back to the retriever, thus completing the loop.

\subsection{Problem definition}
The objective of our model is to predict a single scalar that best estimates the market value \cite{hoeven_appraiser-based_2022} of a property at a given time. To achieve this, the mean squared error of the predicted value $v$ is minimized on the training set $S$. The predicted value is taken as the natural logarithm of the transaction price (and normalized for EREA), or the transaction price per square meter for the IV dataset (since the surface is available, see \ref{Ille-et-Vilaine}).



We adopt the formulation of REML systems to solve this problem: a system composed of two main parts i.e. the retriever model and the prediction model, or downstream model. Therefore, we will seek to minimize the following empirical risk: 

\begin{equation}
L = \frac{1}{|S|}\Sigma_{t \in S}(f_{\theta}(F_t;R_{\omega_1},...,R_{\omega_N}) - v_t)^2
\end{equation}

The model $f_{\theta}$ parameterized by $\theta$ is called the downstream model and $R_{\omega_i}$ denotes the $i^{th}$ retriever parameterized by $\omega_i$ \cite{zamani}. $F_t$ represents the features of the target property $t$.

\subsection{Model architecture}

\begin{figure}
\includegraphics[width=\textwidth]{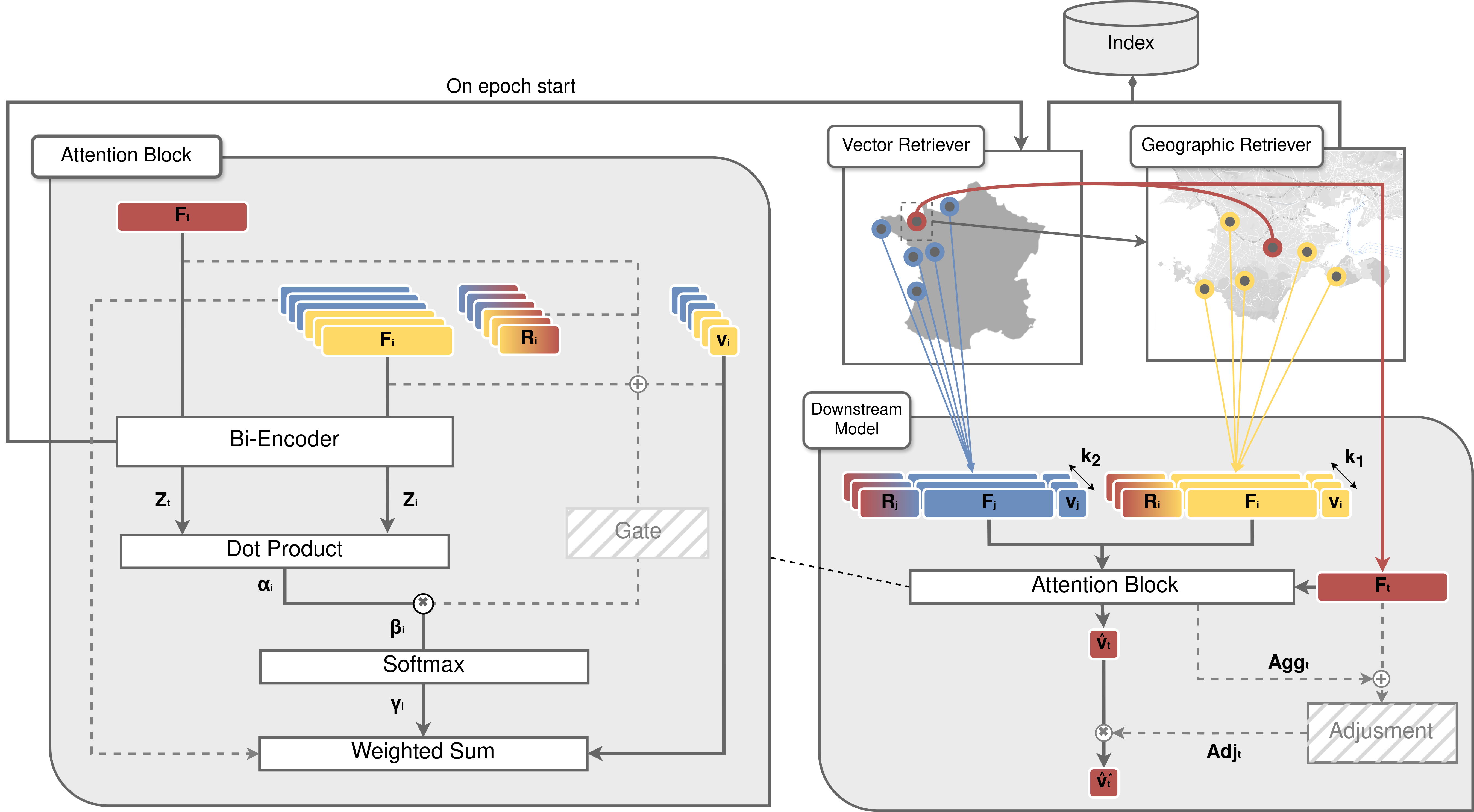}
\caption{\textbf{Model architecture.} Red rows represent the target property whose price is estimated; blue rows represent comparables picked via vector retrieval; yellow rows represent comparables picked via geographic retrieval; bi-color rows represent features relative to the target; grey shaded rectangles and dashed lines are the modules added to REA to form EREA.} \label{fig1}
\end{figure}

\subsubsection{Retrieval}

Our system uses two different top-$k$ retrievers: a geographic retriever $R_1$ which selects $k_1$ comparables exclusively using the haversine distance formula, and a vector retriever $R_2$ which selects $k_2$ comparables based on embedding dot product similarity among the $N$ geographically closest comparables (Fig. \ref{fig1}). The latter acts like a re-ranking mechanism over geographical proximity. For each comparable, we retrieve its features $F_i$, its transaction value $v_i$, and its relative features towards the target (e.g. geographic distance) $R_i$.

\subsubsection{Bi-encoder}

We use a bi-encoder, known to increase robustness \cite{izacard2021unsupervised}, processing both the query (our property target) and comparables.
It consists of a simple MLP using the SELU \cite{selu} activation function.

\begin{equation}
    Zi=BiEncoder(Fi)
\end{equation}

This encoding is common to both retrieval and appraisal prediction.

\subsubsection{Attention block}

We compute a dot product between the target and a comparable encoding to generate a raw attention score.

\begin{equation}
    \alpha_i= Z_i . Z_t
\end{equation}

Note that comparable features do not include the values directly since we do not have access to the target’s value and have to keep dimensions symmetrical. They do not include anything relative to the target either in order to remain usable for retrieval, a setting that requires target agnostic information.

However, these missing features can be included using a gating mechanism on the raw attention score itself. That way, every feature can play a role in the attention without hampering retrieval. Furthermore, it allows differentiation between retrieval selection and model attention scores. Gates are not necessary to the core retrieval framework and thus not included in REA, instead using $\beta_i = \alpha_i$ (\ref{gates}).

We normalize attention scores across all comparables $C_t$ posterior to this operation to guarantee a sum equal to 1 using the softmax operator (\ref{softmax}).

\begin{multicols}{2}
 \noindent
 \begin{equation} \label{gates}
   \beta_i = \alpha_i \times gate(F_t \oplus F_i \oplus R_i \oplus v_i ) 
 \end{equation}
 \begin{equation} \label{softmax}
   \gamma_i = \frac{e^{\beta_i}}{\Sigma_{j \in C_t} e^{\beta_j}} 
 \end{equation}
\end{multicols}

The attention module is completed by a weighted sum over values. This is enough to perform an estimation with the minimal version of our framework (REA). We compute the weighted sum of features too for the extended version of the model.

\begin{multicols}{2}
  \noindent
  \begin{equation}
    \hat{v}_t = \sum_{i \in C_t} \gamma_i v_i
  \end{equation}
  \begin{equation}
    Agg_t = \sum_{i \in C_t} \gamma_i (F_i \oplus R_i \oplus v_i) 
  \end{equation}
\end{multicols}

We chose to use a single encoder and attention layer by mixing comparables from both retrievers after empirically testing them separated \cite{wu2022} or mixed.

Using both types of comparables creates competition dynamics through the attention’s softmax, thus enticing vectors to be competitive with the geographical heuristic. It also helps bring much needed stability to the model, guaranteeing half of the comparables are fixed across all epochs c.f. \nameref{convergence}.

\subsubsection{Adjustment}

Even though the REA module can predict a value using only weighted average, the extended model can benefit from further refinement. We chose to predict a scalar multiplicative factor in order to adjust the raw weighted value by a factor between $[-100\%, + 100\%]$. This factor is the result of a decoder MLP followed by a hyperbolic tangent function.

\begin{multicols}{2}
  \noindent
  \begin{equation}
    Adj_t = tanh(Decoder(Agg_t \oplus F_t))
  \end{equation}
  \begin{equation}
    \hat{v}^*_t = (1 + Adj_t) \times \hat{v}_t 
  \end{equation}
\end{multicols}

The hyperbolic tangent also acts as a security mechanism over the capacity of the model by containing prices into a given range around the aggregated comparable values $\hat{v}_t$. 

\subsubsection{Training procedure}

The training procedure alternates between training epochs and retriever index updates. At the beginning of every new epoch, the entire index is updated with new embeddings generated using the latest model weights. A new sampling is then performed, associating to each target $t$ a new set of comparables $C_t$.  

\subsubsection{Convergence and model stability}
\label{convergence}

Although the training loop makes it possible to learn the retriever and downstream model jointly, it may also cause instability.
The retrieval index is updated every new epoch using the latest model's embeddings. Therefore, the set of drawn comparables differs from one epoch to the other. It causes the model to try and adapt to a slightly new configuration every new epoch. These intra-epoch learning regimes may prevent the model from converging.

To provide the model stability, we use a decay factor on the learning rate of the encoder module. This way, the first epochs are used to learn a good sampling policy, while the latest are used to refine other modules. For the same reason, we also evaluate our models using the previous to last embeddings.

\section{Results}
\subsection{Dataset}

We are evaluating our model on five datasets. Four of these datasets are commonly used in attention-based models for real estate valuation benchmarking \cite{viana2021}, spanning King County (KC) and Fayette County (FC) in the USA, as well as São Paulo (SP) and Porto Alegre (POA) in Brazil. We introduce a new dataset in this article spanning the Ille-et-Vilaine (IV) department in Brittany, France. Dataset statistics can be found in Table \ref{tab_dataset}.

\begin{table}
\caption{
Dataset statistics. Values and distances are calculated w.r.t the 40 geographically closest comparables.}\label{tab_dataset}
\centering
\begin{tabular}{ |l|l|l|l|l|l|l|l|}
    \hline
    Region & Attr. & Samples & Value (\euro/$m^2$, \$, R\$) & Dist. (m) \\
    \hline
    IV & 22 & 45,554 & 2,613 \tiny{$\pm$ 1,278} & 353 \tiny{$\pm$ 628} \\
    \hline
    KC & 19 & 21,608 & 540,098 \tiny{$\pm$ 367,156} & 604 \tiny{$\pm$ 940} \\
    \hline
    FC & 12 & 83,067 & 155,288 \tiny{$\pm$ 76,420} & 98 \tiny{$\pm$ 148} \\
    \hline
    POA & 8 & 15,368 & 443,798 \tiny{$\pm$ 228,518} & 202 \tiny{$\pm$ 268} \\
    \hline
    SP & 8 & 68,848 & 741,952 \tiny{$\pm$ 411,643} & 202 \tiny{$\pm$ 246} \\
    \hline
\end{tabular}
\end{table}

\vspace{-2em}

\subsubsection{Ille-et-Vilaine}
\label{Ille-et-Vilaine}

Ille-et-Vilaine is a diverse area with one main city, Rennes, a coastal hub, Saint-Malo, and relatively sparse countryside. The dataset is extracted from french open data, and comprises transactions including their price, date, location and main structural information such as surface, property type, and room count. It also comes with point of interests counts extracted from OpenStreetMap (OSM) and distances to key anchors like job attraction poles and coastlines. The dataset spans 4 and a half years of data, from January 2016 to June 2023, i.e. a much longer span than other datasets used in this study.

\subsubsection{Temporal retrieval}
\begin{figure}
\includegraphics[width=\textwidth]{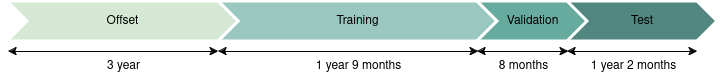}
\caption{The temporal split of the IV dataset into offset, train, validation, test.} \label{fig2}
\end{figure}

When transaction dates are available (for the IV dataset only), we split the dataset chronologically (Fig. \ref{fig2}) to prevent data leakage, and guarantee that during testing, comparables from a posterior date cannot be drawn to help predict the price of a past transaction. Real use-cases cannot perform retrieval in the future, thus evaluation should reflect this limitation. An offset of 3 year is left outside of training to guarantee that even early transactions from the training set will find relevant comparables among past transactions.

\vspace{-2em}

\subsection{Experimental Setup}

REA and its extension EREA have been developed using Python 3.11 and PyTorch 2.3 as the backend. The ASI model was used directly from the original article's GitHub repo \cite{viana2021} for state-of-the-art datasets, and adapted into PyTorch for the IV dataset.

\subsubsection{Framework Hyperparameters}

In the following experiments, we configure REA and EREA to retrieve $2k_1$ comparables in total, either entirely with geographic retrieval, entirely with vector retrieval, or half and half. The vector comparables are taken from the closest $N = 3 k_1 + 25 $ geographic comparables. This affine function was chosen as the intercept constant avoids vector search being limited to single-digit closest comparables at low $k_1$, while the small multiplicative factor avoids vector search being swarmed by too many far and potentially irrelevant comparables at high $k_1$.

The Adam optimizer is used with a learning rate of $10^{-3}$ during training, although the decay factor of 0.98 reduces the bi-encoder rate every epochs. The model is trained over 50 epochs with a batch size of 64. We use a standard scaler for input features of our model.

REA hyper-parameters for Table \ref{tab1} are chosen per dataset from the retrieval method and number of comparables resulting in the best average validation MdAE in Fig. \ref{fig3}, whereas EREA hyper-parameters have been instead manually optimized solely for performance.

\subsubsection{Metrics}

We will consider both extrinsic and intrinsic evaluation \cite{zamani}, that is evaluation of the final estimation and of the quality of retrieved comparables themselves respectively.

We use different metrics for comparison. We adopted median absolute error (MdAE) as a standard for regression evaluation. A traditionally used metric is median absolute percentage error (MdAPE), which we replace by the median absolute balanced relative error (MdABRE) \cite{miyazaki1991method} benefiting from symmetry between over- and under-predictions.

\begin{equation}
ABRE(x,y) = \frac{|x - y|}{min(x,y)}
\end{equation}

\subsection{Intrinsic retrieval evaluation}

\begin{figure}
\includegraphics[width=1\textwidth]{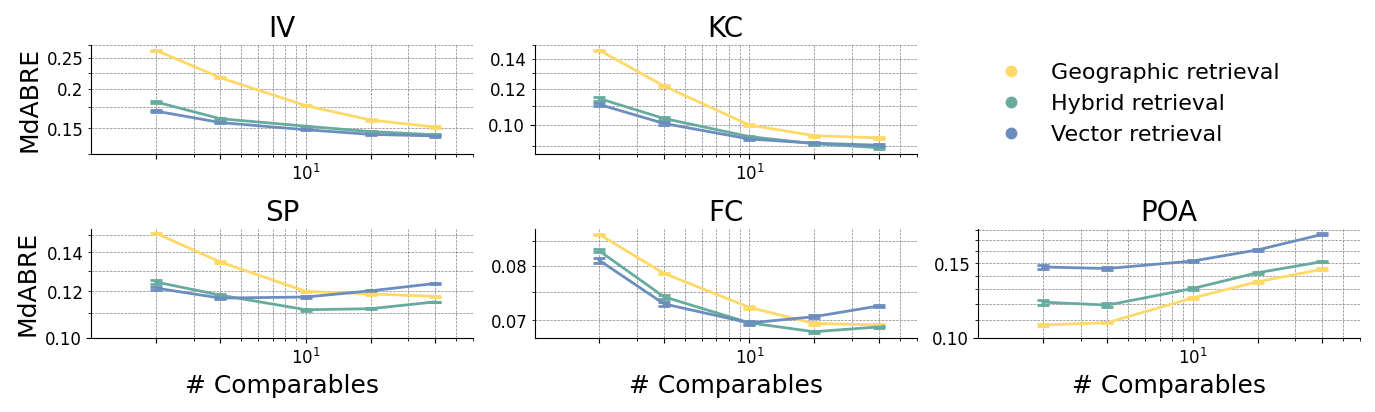}
\caption{\textbf{Intrinsic retrieval evaluation with REA.} Geographic retrieval (yellow), hybrid retrieval (green), and vector retrieval (blue), are compared with different number of comparables. The average and 95\% CI of 20 simulations are shown here.} \label{fig3}
\end{figure}

In Figure~\ref{fig3}, we show the model performance when varying the number of retrieved comparables using (i) the usual geographic heuristic, (ii) our approach and (iii) a combination of the two. Our REA model significantly outperforms geographic retrieval at lower number of comparables on 4 out of 5 datasets. This demonstrates that vector search retrieves higher-quality comparables compared to geographic retrieval, the low number of high-quality comparables allowing for easier interpretation of results.

Since both retrieval methods are synergistic, we could have expected hybrid search to yield better performance, yet this observation isn't systematic across different number of comparables. Nonetheless, hybrid search outperforms vector search performance when a large number of comparables are selected, possibly by stabilizing selected comparables during training.

Whereas the vector retrieval selects higher-quality comparables on 4 datasets, the tendency is completely reversed for the POA dataset, an increasing number of comparables worsening performance. This is due to the high amount of redundancy in the dataset, where 30.5\% of properties share the exact same price (absolute difference < 1R\$) as the closest comparable. In comparison, this percentage is lower in other datasets: 12.74\% for SP, 1\% for KC and 5.76\% for FC (< 1\$), and 1.08\% for IV (< 1\euro).

\subsection{Baselines}

We will be evaluating the performance of our model against traditional machine learning baselines \cite{viana2021} (Linear Regression (LR), XGBoost (XGB)), a locality-based baseline ($k$-Nearest Neighbor (KNN)), and the state-of-the-art (Attention-Based Spatial Interpolation (ASI)).

Table \ref{tab1} shows that ASI outperforms on KC, FC and SP datasets, and EREA outperforms on IV and POA datasets. More generally, EREA shows relatively similar performance to ASI for only a fraction of the parameters, as much as $\sim22\times$ when comparing the 17,140 parameters of EREA and 792 parameters of ASI on the FC dataset (Table \ref{tab2}). This reduced number of parameters shows the advantage of high-quality comparable retrieval, the model simplicity moreover allowing for more easily explainable model decisions.

\begin{table}
\caption{
Baseline comparison table on all 5 datasets. MdAE is in \euro/$m^{2}$ for IV, \$\ for KC/FC and  R\$ for POA/SP, MdABRE is in \%. The average and 95\% CI of 20 simulations are shown here for the last 3 models. ASI was adapted into PyTorch for the IV dataset and might differ slightly, hence the asterisk.}\label{tab1}
\resizebox{\textwidth}{!}{
\begin{tabular}{ |l|l|l|l|l|l|l|l|l|l|l| }
    \hline

    Model & \multicolumn{2}{|c|}{IV} & \multicolumn{2}{|c|}{KC} & \multicolumn{2}{|c|}{FC} & \multicolumn{2}{|c|}{POA} & \multicolumn{2}{|c|}{SP} \\
    \hline

    & {\scriptsize MdAE} & {\scriptsize MdABRE} & {\scriptsize MdAE} & {\scriptsize MdABRE} & {\scriptsize MdAE} & {\scriptsize MdABRE} & {\scriptsize MdAE} & {\scriptsize MdABRE} & {\scriptsize MdAE} & {\scriptsize MdABRE} \\
    \hline

    LR & 574 & 21.21 &  68568 & 16.96&  20259 & 16.66 &  82831 & 25.57 &  136263 & 25.70 \\
    \hline

    kNN &  467 & 19.24 & 68112 & 16.83 &  12255 & 9.55 &  40000 & 11.18 &  108381 & 19.73 \\
    \hline

    XGB &  408 & 16.08 &  39485 & 9.26 &  9990 & 7.72 &  44548 & 12.41 &  69212 & 11.93  \\
    \hline
    \hline


    \multirow{ 2}{*}{ASI}
    & 357.16*            & 13.24*            & 36559.8            & 8.37              & 8608.5            & 6.55              & 34281.1            & 9.57             & 58194.7            & 10.0 \\[-4pt]
    & \tiny{$\pm$ 2.75}  & \tiny{$\pm$ 0.09} & \tiny{$\pm$ 224.3} & \tiny{$\pm$ 0.05} & \tiny{$\pm$ 40.4} & \tiny{$\pm$ 0.03} & \tiny{$\pm$ 408.2} & \tiny{$\pm$ 0.1} & \tiny{$\pm$ 1496.1} & \tiny{$\pm$ 0.24} \\
    \hline

    \multirow{ 2}{*}{REA} & 381.0 & 14.21 & 38571.1 & 8.90 & 8864.1 & 6.81 & 39216.1 & 10.71 & 64988.5 & 11.16 \\[-4pt]
    & \tiny{$\pm$ 1.4} & \tiny{$\pm$ 0.07} & \tiny{$\pm$ 220.5} & \tiny{$\pm$ 0.05} & \tiny{$\pm$ 25.4} & \tiny{$\pm$ 0.02} & \tiny{$\pm$ 277.9} & \tiny{$\pm$ 0.06} & \tiny{$\pm$ 219.8} & \tiny{$\pm$ 0.04} \\
    \hline

    \multirow{ 2}{*}{EREA} & 354.26 & 13.13 & 38253.85 & 8.80 & 8736.29 & 6.72 & 33056.22 & 8.91 & 58182.85 & 10.12 \\[-4pt]
    & \tiny{$\pm$ 2.43} & \tiny{$\pm$ 0.10} & \tiny{$\pm$ 259.65} & \tiny{$\pm$ 0.06} & \tiny{$\pm$ 35.61} & \tiny{$\pm$ 0.03} & \tiny{$\pm$ 511.67} & \tiny{$\pm$ 0.14} & \tiny{$\pm$ 361.75} & \tiny{$\pm$ 0.08} \\
    \hline
        
\end{tabular}
}
\end{table}

\vspace{-2em}

\begin{table}
\caption{
Number of parameters per model and dataset.}\label{tab2}
\centering
\begin{tabular}{ |l|l|l|l|l|l|l|}
    \hline
    Model & IV & KC & FC & POA & SP \\
    \hline
    ASI & 11,660 & 21,700 & 17,140 & 19,240 & 3,565 \\
    \hline
    REA & 900 & 800 & 660 & 580 & 580  \\
    \hline
    EREA & 1,472 & 1,142 & 792 & 592 & 592 \\
    \hline
\end{tabular}
\end{table}

\vspace{-2em}

\section{Conclusion}
Real estate appraisal following the comparable estimation methodology can be approached through the prism of Information Retrieval. In this paper, we adapt REML techniques that found recent success to the field of real estate valuation. In particular, we proposed a hybrid retrieval system that improves upon traditional techniques of heuristic-based comparable selection.

In our study, we demonstrated that learning the retrieval policy can find better comparables than traditional methods based on geographical distance. Furthermore, based on this efficient selection of comparables, we have shown that we can build a model with considerably less parameters compared to existing methods with close to equivalent performance, which results in improved explainability and reduced computational requirements.

By showing that the selection of comparables can be significantly improved, this work can inspire the research community to focus on the selection of comparables for their research. Our future work will focus on this aspect, and with high-quality comparables, we are convinced that it will be possible to build models that exceed the performance of existing techniques. Finally, focusing on few high-quality comparables could allow to enrich target and comparables with memory-intensive features such as images for higher performance at lower computational costs.

\section{Acknowledgment}

We would like to thank Muneeb ul Hassan for his support and collaboration on projects that led to the maturation of the present innovative method.

%
%
%
\bibliographystyle{splncs04}
\bibliography{biblio}

\end{document}